# Text Detection and Recognition Based on Lensless Imaging System

Yinger Zhang, Zhouyi Wu, Peiying Lin, Yuting Wu, Lusong Wei, Zhengjie Huang, Jiangtao Huangfu*

*Laboratory of Applied Research on Electromagnetics, Zhejiang University, Hangzhou, China*
*Corresponding author: huangfujt@zju.edu.cn*

**Lensless cameras are characterized by several advantages (e.g., miniaturization, ease of manufacture and low cost) as compared with conventional cameras. However, they have not been extensively employed for their poor image clarity and low image resolution, especially for tasks that have high requirements on image quality and details like text detection and text recognition. To address the problem, a framework of deep learning-based pipeline structure was built to recognize text with three steps from raw data captured by employing lensless cameras. This pipeline structure consisted of the lensless imaging model U-Net, the text detection model CTPN (Connectionist Text Proposal Network), as well as the text recognition model CRNN (Convolutional Recurrent Neural Network). Compared with the method only focusing on the image reconstruction, U-Net in the pipeline was able to supplement the imaging details by enhancing factors related to character categories in the reconstruction process, so the textual information can be more effectively detected and recognized by the CTPN and CRNN with the less artifacts and high clarity reconstructed lensless images. By performing experiments on datasets of different complexity, the applicable for applying text detection and recognition on lensless cameras were verified. It is considered that this study can reasonably demonstrates text detection and recognition task in the lensless camera system, which develops a basic method for novel applications.**

## 1. INTRODUCTION

As imaging processing technologies have been leaping forward over the past few decades, various visual tasks have been employed extensively (e.g., face recognition on the access control system, vehicle inspection in autopilot system and text detection in checkout scanning system). In the mentioned application scenarios, cameras have been always integrated into a specific hardware system. Consider the integration of cameras into drones for distributed sensing in applications such as disaster recovery or in very thin objects such as credit cards. Given the weight and volume constraints in these devices, integration is not possible for lens-based cameras.

Fortunately, lensless computational cameras have been developed as an alternative, which place thin masks other than lens in front of the sensors, thereby significantly decreasing the thickness and weight of camera. Considerable algorithms have been proposed to restore images from the captured coded measurements [1-3]. However, compared with images directly captured by lens-based cameras, the reconstructed images from lensless cameras inevitably have various degradations (e.g., blurring, lower resolution and low signal-to-noise ratio). For this reason, the study of visual tasks on lensless system is much less than that on lens-camera: So far, character recognition [4,32], face recognition [5,32], action recognition [30,31] and other classification tasks [33] have been studied on lensless system, but no discussion has been conducted on whether the quality and resolution of images from lensless camera meet text detection and recognition tasks.

Accordingly, this study explored the effects of text detection and recognition on the lensless imaging system by adopting the method of artificial intelligence. A deep learning-based pipeline structure was proposed, which could be used to achieve text detection and recognition results from the lensless raw data. In the pipeline, lensless imaging model U-Net reconstructed images from raw data captured in lensless camera first, and then the text detection model CTPN extracted text regions in the reconstructed lensless images. Lastly, the text recognition model CRNN translated those image-based sequence into text. The present study proved that text detection and text recognition tasks could achieve reasonable results in the lensless camera system. To the authors' knowledge, this is one of the first work to study text detection and text recognition on a lensless imaging system. Compared with the method only focusing on the image reconstruction, the proposed lensless imaging model was able to supplement the imaging details by enhancing factors related to character categories in the reconstruction process. Meanwhile, the textual information can be more effectively detected and recognized by the CTPN and CRNN with the less artifacts and high clarity reconstructed lensless images, thus promoting the effect of lensless text detection and recognition. Additionally, various factors were further explored to clarify the application condition. The work probably leads to various novel applications, particularly in scenes with requirements for privacy protection and devices thickness (e.g., product identification in industry and identity information extraction).

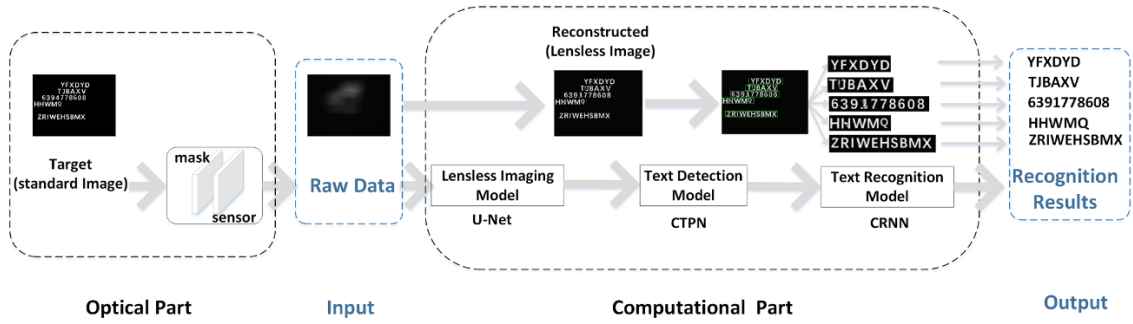

**Fig. 1.** The framework of this study

## 2. RELATED WORK

### A. Lensless Imaging

Over the past few years, with the development of computational imaging, several small factor-form lensless cameras have been presented. [6] proposed the FlatCam, in which the mask was a binary amplitude mask. [7,34] employed a Multi-phased Fresnel zone aperture for imaging. Patrick R. Gill et al. [8,9] invented an ultraminiature computational imager by employing special optical phase gratings integrated with CMOS photodetector matrixes. DiffuserCam adopted a phase diffuser to achieve 3D lensless imaging [1,10]. In this study, DiffuserCam acted as the optic hardware system, in which the lens was replaced with a phase mask placed a short distance in front of the bare sensor.

Other than directly recording images (e.g., lensed cameras), lensless cameras encode the object information into sensor measurements, thereby requiring restoration algorithm to generate the desired image. Lensless image reconstruction refers to an ill-conditioned inverse problem, and the methods fall into three general types, i.e., iterative optimization method, which use convex optimization like gradient descent (GD) and alternating direction methods of multipliers (ADMM) iteratively [11]; pure end-to-end convolutional neural network [12-14,35]; and unrolled optimizer, addressing the inverse problem by incorporating the system model into the network [15]. In this study, pure end-to-end CNN method was adopted to perform the image reconstruction, and the detailed reasons are discussed in Section 3.1.

### B. Text Detection and Recognition

Reading text in image consists of two sub tasks, i.e., text detection and text recognition. Text Detection is a task to detect text regions in the non-textual background and label them with bounding boxes [16]. In this study, we utilized a model based on CTPN [21], an extensively used network in the field of text detection, for text detection. Text recognition aims to translate an image-based sequence which containing only text part into text. In this study, we utilized a model based on CRNN [22], a network architecture which consists of convolutional layers, the recurrent layers, and a transcription layer for text detection.

The above works concentrate on text detection and recognition for images captured by employing lens-based camera and have achieved satisfactory results, so we also use these methods to study the effect of lensless imaging technology in text detection and recognition tasks.

## 3. METHOD

The framework here is illustrated in Fig. 1. Optical part referred to the lensless camera system where a thin mask placed several millimeters in front of a sensor, and it encoded the target information into raw data. Computational part consisted of lensless imaging model, text detection model and text recognition model. The mentioned three models could be trained separately in the training phase, whereas in the inference phase, they were integrated as a pipeline structure through which the text detection and recognition results could be achieved from the lensless raw data directly. Lensless imaging model restored images from raw data, and then text detection model extracted text regions from the non-textual background in the reconstructed lensless images. Lastly, text recognition model translated those image-based sequence into text. In Section 3.1, the lensless camera system and the imaging model were introduced in this study. Subsequently, in Section 3.2, the process to obtain of lensless text scene image dataset is described. The text detection and recognition models are explained in Section 3.3.

### A. Lensless Imaging

In this study, a lensless camera system was built by referencing the hardware prototype of DiffuserCam as the lensless camera system. According to Fig. 2, the lensless camera consisted of a diffuser (luminit 1°) with an aperture of 1/4 sensor size placed 2mm away from a sensor (ov5647, 5-megapixel), and 3D printing bracket was applied to keep the diffuser at the right distance. Target images (physical size of 13cm*17.3cm) were showed on a 1080p LED monitor placed at 20cm in front of the camera to match the field-of-view of the lensless camera.

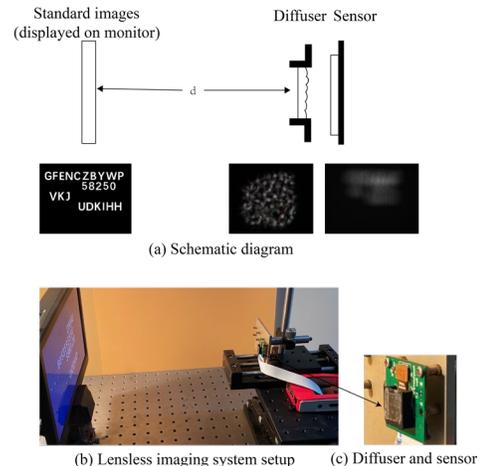

**Fig. 2.** Experiment schematic and devices

![U-Net architecture diagram]

**Fig. 3.** The architecture of the lensless imaging model

*1. Lensless Imaging Forward Model*

Light from a point source in the scene was refracted by the diffuser to create a high-contrast pattern on the sensor, i.e., the point spread function (PSF) of this system. The following Eq. (1) serves as the lensless imaging system forward model [1,23] by assuming that all points in the scene are incoherent and shift invariant:

$$\mathbf{b}(x,y) = crop[\mathbf{h}(x,y) * \mathbf{x}(x,y)]$$
$$= \mathbf{CHx} \qquad (1)$$

Where $\mathbf{b}$ denotes the sensor measurement; $\mathbf{h}$ represents the system PSF, which can be estimated through calibration; $\mathbf{x}$ represents target scene; $(x,y)$ is the coordinates; $*$ denotes 2D convolution. Given the finite size of the sensor, a crop operation was applied to the model. $\mathbf{C}$ and $\mathbf{H}$ express crop operation and PSF, respectively, under matrix-vector notation.

*2. Inverse Algorithm*

In the work, pure end-to-end CNN method was selected to perform the image reconstruction. U-Net [24] refers to a convolution neural network that contains down-sampling architecture and up-sampling architecture connected through skip-connection. The overview of the network structure is illustrated in Fig. 3. The input of U-Net was the raw data recorded from the sensor and was downsized to (243,324) pixels, while the corresponding label image was that displayed on the screen of the identical size (243,324). The output of U-Net was the reconstructed image. During the training process, the network parameters were updated based on MSE loss. In this study, U-Net used Adam optimizer with learning rate of 1e-3, $\beta1 = 0.9, \beta2 = 0.99$ , a weight_decay of 1e-4, and a mini-batch size of 32. The training was implemented on one RTX 2080Ti GPU, with pytorch1.0.0 and python 3.6 under Ubuntu18.04.

The training dataset of lensless imaging model was only involved in the training of U-Net, while the label images and reconstructed images of test dataset moved on to the subsequent text detection and recognition experiments.

The proposed network improved the quality of reconstruction for textual images because of its ability to learn factors related to character categories in the reconstruction process, so the advantage benefits from the combination of reconstruction method and scene (textual dataset). It is able to break through the limitations of light and resolution, and to supplement the reconstructed patterns according to the predicted categories. The textual information can be more effectively detected and recognized by the CTPN and CRNN with the less artifacts and high clarity reconstructed lensless images. The above conclusion can be drawn from the following two experiments:

1. As shown in Fig.4, images reconstructed by U-Net show less artifacts, more uniform brightness and more clear details compared to that of ADMM. Take character "G" in Size40 column as an example, the reconstruction result of ADMM shows that the information in raw data is not enough to recover the details of "G". In contrast, the proposed method restores "G" completely, indicating the details of "G" are supplemented by the judgment of the category "G".

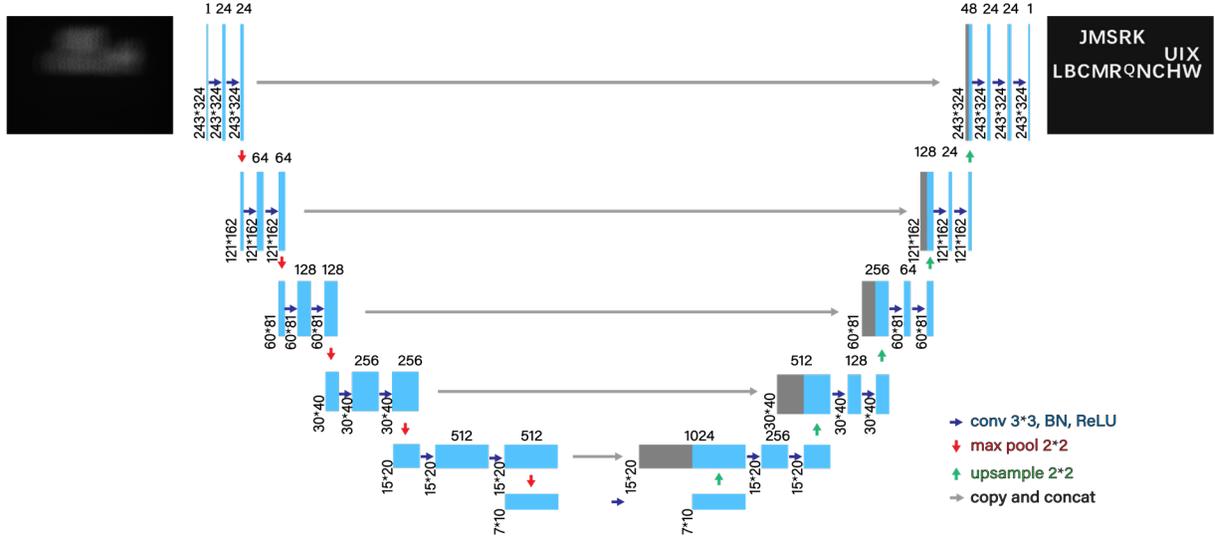

**Fig. 4.** Reconstruction results of images by ADMM method and CNN method

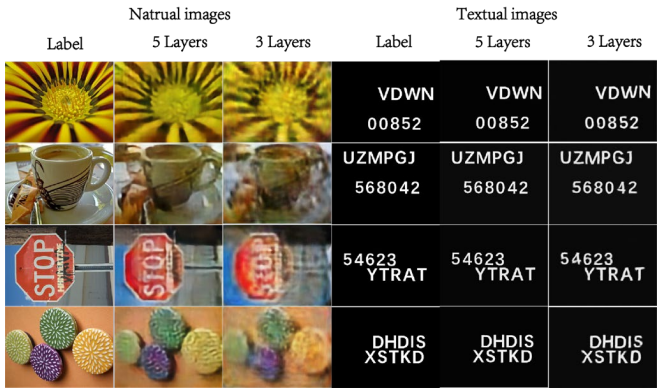

**Fig. 5**. Reconstruction results by U-Net for natural images and textual images

2. Fig.5 compares the reconstruction results of natural images and textual images, natural images are from MirFlickr dataset [36]. It shows that the reconstructed textual images have sharper and more accurate details than the reconstructed natural images. When the number of U-Net layers reduces to 3 layers (number of parameters 706481) from 5 layers (number of parameters 11771825), the quality of the reconstructed natural image is obviously reduced, but the reconstructed textual images still maintain good clarity. These observations illustrate that when compared with natural images, the reconstruction of textual images can obtain higher quality because of the introduction of character classification as an additional factor. It also shows that the reconstruction of text images can be realized using a simple network.

### B. Obtaining Lensless Text Image Dataset

Since there exists no text detection/recognition dataset of lensless camera images, we collected lensless text image dataset using the lensless imaging device mentioned in section 3.1. To facilitate fair comparison, the target image displayed on the monitor had a uniform resolution of (243*324) pixels, physical size of (13.0cm*17.3cm). The image displayed on the monitor was termed the "standard image", and the image reconstructed from raw data was termed the "lensless image".

Two types of test datasets were used for assessment. All datasets only considered words in horizontal direction for simplicity: 1. An artificial dataset named NCD (Normalize character dataset); 2. Common benchmark dataset named IIIT5k (IIIT 5k-word) [25]. The mentioned two test datasets represent scenes of different complexity, so the capability of lensless imaging system in text detection and recognition tasks could be explored in different scenes. Moreover, this can make the work more universal, thereby laying a foundation for later practical application.

For the respective test dataset, a corresponding training set was used to train the lensless imaging model first. Subsequently, "lensless images" of the test datasets were generated by the trained model, on which the follow-up text detection and recognition assessment was conducted. The summary of datasets is listed in Table 1. Examples of the dataset are presented in Fig. 6.

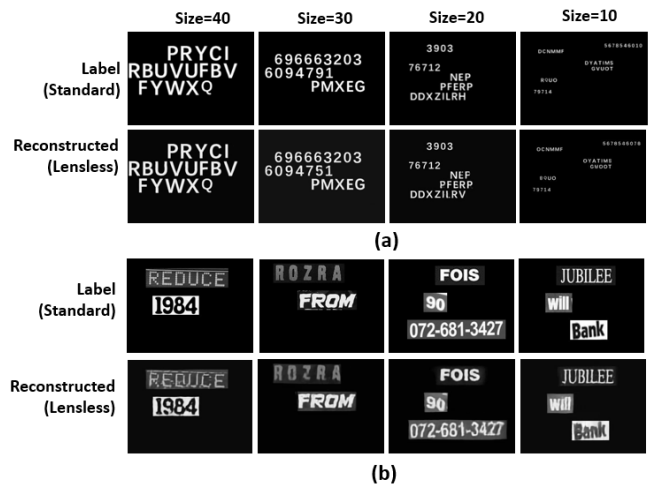

**Fig. 6.** Example images of the two datasets: (a) NCD; (b) IIIT5K. Images in top row are labels, and images in bottom row are the reconstructed images

NCD refers to a synthetic dataset, the images in which had pure black background, the characters involved 26 English letters(A-Z) and 10 Arabic numerals (0-9), and the font of the characters was Arial. NCD contained four sub-datasets and the character size of them were different. The character height =40/30/20/10 pixels, and this study termed the sub-dataset NCD_40, NCD_30, NCD_20, NCD_10, respectively. For the image in the sub-dataset, the number, position, length, and characters of the words were randomly selected. Each sub-NCD had 20000 images, of which 160000 were used as the train set (NCD_TRAIN) for the lensless imaging model, and the remaining 4000 acted as the test dataset (NCD_TEST).

IIIT5k refers to a popular benchmark for scene text recognition, including 3000 cropped word images collected from the Internet. Compared with NCD where all words were in white and written on black background, images in III5K were captured under variables of condition (i.e., lighting, background, camera quality, blur, character shape, font and distortion). For this reason, the complexity of IIIT5K exceeded that of NCD. This study ignored images that either contained non-alphanumeric characters or had the height less than 30 pixels, and the cropped word images were reshaped to a uniform height (50 pixels). Finally, a set of 1857 cropped images were obtained. The cropped images in set were randomly selected without repetition and combined into a synthetic image of size (243,324) pixels. The position and number of the cropped images in synthetic image were random, and eventually 976 synthetic images were generated as test dataset.

**Table 1.  The summary of datasets NCD and IIIT5K**

| Test set (number) | Training set for lensless imaging model (number) | Detect | Recognize | Complexity |
|---|---|---|---|---|
| NCD_TEST (4000) | NCD_TRAIN (16000) | √ | √ | Low |
| III5K (1857) | Synth90k (20000) | √ | √ | High |

20000 images were randomly selected from synth90k [26], and the above operation was performed to obtain train set of lensless imaging model for III5TK, since they exhibited similar image characteristic. To be consistent with the training set, images in IIIT5K were converted into grayscale.

**C. Algorithm for Text Detection and Text Recognition**

As the generic structure of pipeline presented in Fig. 1, after lensless images were obtained through the trained lensless imaging model, text detection and recognition experiments were conducted on them. In this step, algorithms in common use and have achieved good performance were selected as reference, which could also be replaced with any other models with the identical function.

*1. Text Detection*

A deep learning technique was used based on CTPN [21] for text detection, which could accurately localize text lines in natural image. CTPN developed fixed-width proposals by densely sliding a 3*3 spatial window through the last convolutional maps of the VGG16 model. Subsequently, the sequential proposals in each row were naturally connected by a Bi-directional LSTM. Lastly, the RNN layer was connected to a fully-connected layer, which jointly predicted location, text/non-text score and side-refinement offsets of each fixed-width proposal. The code we used for text detection on lensless images was based on [27]

Model was trained on 3,000 natural images (mlt), including 229 images from the ICDAR 2013 training set [21]. Images in the training set were standard images collected by lens-based camera rather than lensless camera. The assessment protocols used here were provided by robust reading competition [28], including three standard metrics, i.e., Recall, Precision and F-score.

*2. Text Recognition*

A deep learning technique was used based on Convolutional Recurrent Neural Network (CRNN) [22] for text recognition, in which text sequence from the cropped word image was predicted. The network architecture consisted of three parts, convolutional layers, extracting a feature sequence from the input image, recurrent layers, predicting a label distribution for each frame, as well as transcription layer, translating the per-frame predictions into the final label sequence. The code applied here complied with [29], and the model was trained on the synthetic dataset (Synth 80k) released by Jaderberg et al. [26], containing 8 million training images (standard images) and their corresponding ground truth words. The same as text detection part, images in training set for CRNN were also collected by lens-based camera rather than lensless camera. This study applied the trained CRNN model tested on testing datasets mentioned in section3.2. Edit distance (string S by adding, deleting, replacing three editing operations, converted to string T, the minimum number of editing required) and Crwr (rate of correctly recognized words) were adopted to assess the performance of text recognition. The assessment code was provided by robust reading competition [28].

**4.EXPERIMENT**

According to the method presented in Section 3.2, an investigation was conducted on the capabilities exhibited by the lensless imaging system for detecting and recognizing text on NCD and IIIT5K. The variation of optical conditions was also tested in the Section.

**A. Experiments on NCD**

The text detection and text recognition experiments were performed on NCD that presented relative ideal conditions.

*1. Text Detection*

For text detection, NCD_TEST was tested on a trained CTPN model (trained on standard mlt). Table 2 lists the Precision, Recall and F-score of lensless images with different text sizes. It could be suggested that lensless images composed of relatively simple elements were perfect and effective in the text detection task. With the decrease in the text image size, the Precision and the Recall remained 1, though there was a slight drop at size=10 which could be nearly ignored. The samples of bounding boxes predicted by CTPN are illustrated in Fig. 7.

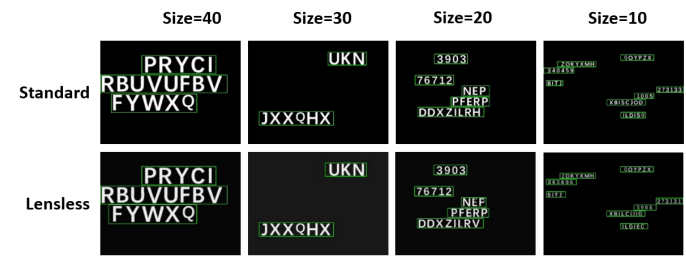

**Fig. 7.** Bounding boxes detected by CTPN on label image (top) and its lensless counterpart (bottom)

**Table 2. Assessment for different text sizes of NCD_TEST (lensless)**

|  | Precision | Recall | F-score |
| --- | --- | --- | --- |
| Size40 | 1.0000 | 1.0000 | 1.0000 |
| Size30 | 1.0000 | 1.0000 | 1.0000 |
| Size20 | 1.0000 | 1.0000 | 1.0000 |
| Size10 | 0.9991 | 1.0000 | 0.9996 |

*2. Text Recognition*

The CRNN model applied for text recognition was trained on standard synth90k. In the following experiment, a comparison was drawn for the performance of lensless image and standard image on text recognition, and the effect of word length and text image size on the accuracy of text recognition was studied. Table 3, Table 4, Fig. 8 present the results of Crwr and Edit distance (we use average Edit distance= Edit distance/number of words). In Fig. 8, the line of standard image represents the ability of the CRNN model, so the delta denotes the recognition error attributed to lensless imaging. From the results, the following insights were gained:

1. Compared with standard images, there was a performance decrease when lensless images were applied, as reflected by lower Crwr and higher Edit distance. However, in several "easy" cases (e.g., the text of large scale (size>=30 pixels) and in short word length (<=7)), the lensless images yielded about only less than 5% decrease in correct word recognize accuracy compared with standard images, which still reached reasonable accuracy for many applications. Furthermore, the CRNN model was trained on standard images without additional collection of lensless images for training, thereby convenient for model migration.

2. It was suggested that the performance of text recognition could be sensitive to the text image size. First, when size>=20 pixels, Crwr and Edit distance of standard images remained basically unchanged at different sizes, thereby indicating that the ability of the text recognition model remained stable when size>=20 pixels, whereas it declined when size<20 pixels. Second, as size decreased, the Crwr of lensless images decreased, the Edit distance of lensless images increased, and the gap between lensless images and standard images increased. The mentioned variations were relatively gentle when size>20 pixels, whereas they were severe under size<=20 pixels. This was as expected since the quality of lensless image reconstruction decreased with the decrease in size, which was correlated with the resolution limitation of the lensless camera system. Third, the poor performance of lensless images in Fig8. (d) and Fig8. (h) originated from the errors of the CRNN model and lensless reconstruction.

3. Furthermore, the effect of the word length was studied. It was revealed that the performance gap of lensless image and standard image expanded as the word length grew, whereas the expansion was insignificant under the large text image size. The magnification of gap was because that the complexity of the image intensified with the growth of word length, thereby increasing the reconstruction error in the lensless image reconstruction process.

**B. Experiments on Standard OCR Dataset**

Experiments were performed on IIIT 5K, a popular benchmark for text detection and recognition. It could be more consistent with the practical use scene as compared with NCD. The method to obtain standard/lensless version of the dataset is elucidated in Section 3.2.

After the processing presented in section 3.2, 1857 cropped images remained, which were used for test dataset synthesis. Compared with NCD, IIIT 5K covers considerable variations in fonts and background, and some of these variations (i.e., (a) thin line text (b) complex background (c) low contrast for text and background (d) irregular text) increase the difficulty in text recognition in lensless images. The mentioned variations usually cause poor image reconstruction quality. The bad performance of (a) and (c) is attributed to the limitation of the lensless system, thereby causing limited resolution and low dynamic sensing range. The poor image quality of (b) and (d) results from the lack of relevant variations in training set for the lensless imaging model. On that basis, two datasets were created: the "simple" dataset excluded cropped images containing the above variations, and 1272 cropped images remained. The "complex" test dataset covers all cropped images. Fig. 9 presents the examples of cropped images for the "simple" and "complex" datasets.

*1. Text Detection on IIIT5K*

The synthetic images were generated for "simple" and "complex" datasets, respectively, and a dataset of 715 images was obtained for the "simple" dataset and a dataset of 976 images for "complex". Table 5 lists the Precision, Recall and F-score of lensless/standard images for "simple" and "complex" datasets. As indicated from the table, compared with standard images, the performance of lensless images in text detection declined slightly, whereas it continued to be perfect and effective for task.

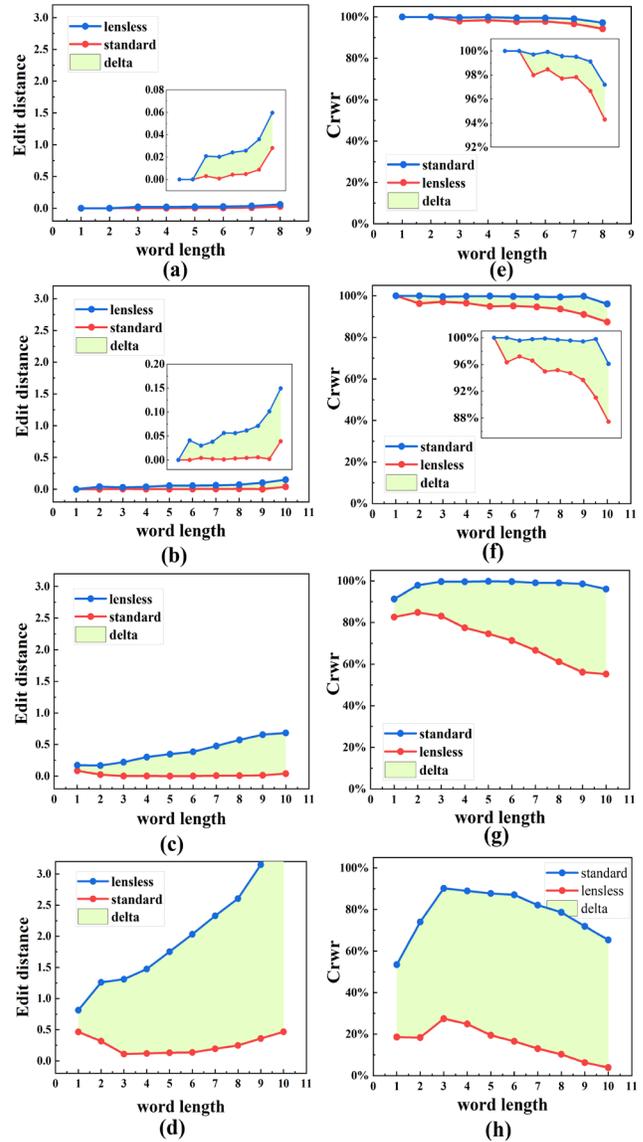

**Fig. 8.** Crwr versus word length on the NCD dataset for different text sizes: (a) text size=40 pixels; (b) text size=30 pixels; (c) text size=20 pixels; (d) text size=10 pixels. Average Edit distance versus word length on the NCD dataset for different text size: (e) text size=40 pixels; (f) text size=30 pixels; (g) text size=20 pixels; (h) text size=10 pixels.

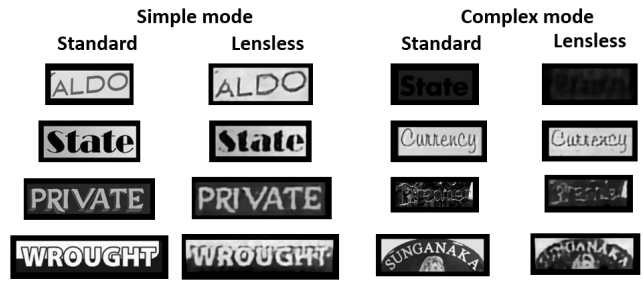

**Fig. 9.** Sample cropped images for the "simple" mode and the "complex" mode. Images in the left column are standard images, and images in the right column are the lensless images.

**Table 3.** Crwr performance on the NCD dataset of standard images and lensless images. The effect of the word length and text size was explored.

| word length | size40 | | size30 | | size20 | | size10 | |
|---|---|---|---|---|---|---|---|---|
| | standard | lensless | standard | lensless | standard | lensless | standard | lensless |
| 1 | 100.00% | 100.00% | 100.00% | 100.00% | 91.30% | 82.61% | 53.49% | 18.60% |
| 2 | 100.00% | 100.00% | 100.00% | 96.33% | 97.89% | 84.86% | 74.05% | 18.37% |
| 3 | 99.70% | 97.99% | 99.58% | 97.20% | 99.68% | 83.08% | 90.20% | 27.44% |
| 4 | 99.92% | 98.46% | 99.79% | 96.58% | 99.63% | 77.48% | 88.96% | 24.90% |
| 5 | 99.57% | 97.71% | 99.90% | 94.96% | 99.84% | 74.57% | 87.73% | 19.45% |
| 6 | 99.52% | 97.83% | 99.70% | 95.16% | 99.68% | 71.32% | 87.08% | 16.57% |
| 7 | 99.12% | 96.67% | 99.58% | 94.70% | 99.08% | 66.64% | 82.12% | 13.01% |
| 8 | 97.20% | 94.30% | 99.45% | 93.68% | 99.10% | 61.14% | 78.69% | 10.31% |
| 9 | None | None | 99.80% | 91.05% | 98.54% | 56.15% | 71.94% | 6.27% |
| 10 | None | None | 96.10% | 87.45% | 96.05% | 55.24% | 65.36% | 3.91% |

**Table 4.** Edit distance performance (Edit distance/Number of words) on the NCD dataset of standard images and lensless images. The effect of the word length and text size was explored.

| word length | size40 | | size30 | | size20 | | size10 | |
|---|---|---|---|---|---|---|---|---|
| | standard | lensless | standard | lensless | standard | lensless | standard | lensless |
| 1 | 0/12 | 0/12 | 0/27 | 0/27 | 2/23 | 4/23 | 20/43 | 35/43 |
| 2 | 0/226 | 0/226 | 0/245 | 10/245 | 7/284 | 48/284 | 167/528 | 666/528 |
| 3 | 4/1344 | 28/1344 | 7/1677 | 50/1677 | 7/2157 | 480/2157 | 386/3509 | 4598/3509 |
| 4 | 1/1236 | 25/1236 | 3/1402 | 53/1402 | 7/1874 | 568/1874 | 357/2980 | 4395/2980 |
| 5 | 7/1613 | 39/1613 | 2/1925 | 108/1925 | 4/2493 | 872/2493 | 543/4133 | 7245/4133 |
| 6 | 6/1242 | 32/1242 | 4/1342 | 75/1342 | 6/1893 | 732/1893 | 413/3041 | 6183/3041 |
| 7 | 10/1142 | 41/1142 | 4/944 | 58/944 | 11/1193 | 571/1193 | 386/1991 | 4636/1991 |
| 8 | 32/1141 | 68/1141 | 5/902 | 64/902 | 11/1225 | 704/1225 | 504/2027 | 5281/2027 |
| 9 | None | None | 1/503 | 51/503 | 9/618 | 406/618 | 349/973 | 3064/973 |
| 10 | None | None | 18/462 | 69/462 | 27/659 | 451/659 | 501/1074 | 4051/1074 |

whereas it continued to be perfect and effective for task. Moreover, as compared with the simple dataset, the assessment indicators declined in the "complex" dataset, whereas it could still meet numerous scenarios. Samples of bounding boxes predicted by CTPN are illustrated in Fig. 10.

*2. Text Recognition on IIIT5K*

Experiments were still performed on the "simple" and "complex" datasets, respectively. To avoid errors attributed to text detection, the text images were cropped from the synthesis image by employing bounding boxes information in label.

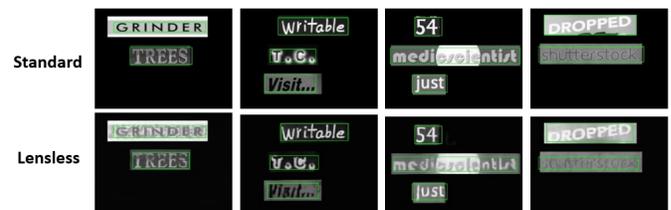

**Fig. 10.** Bounding boxes detected by CTPN on label image (top) and its lensless counterpart (bottom)

**Table 5. Assessment for different complexities of IIIT5K (standard and lensless)**

| Dataset | | Total Num | Precision | Recall | F-score |
|---|---|---|---|---|---|
| simple | Standard | 715 | 0.8737 | 0.8769 | 0.8753 |
| | Lensless | 715 | 0.8574 | 0.8283 | 0.8426 |
| complex | Standard | 976 | 0.8599 | 0.8640 | 0.8620 |
| | Lensless | 976 | 0.8337 | 0.7704 | 0.8008 |

The results are listed in Table 6. Consistent with NCD, there was a performance decrease when using lensless images instead of standard images. In the "easy" dataset, however, the Crwr of lensless could reach 71.78%, and the gap to standard image was only 17.02%, which remained sufficient in numerous scenarios.

Furthermore, the dramatic difference between "simple" and "complex" dataset reveals the requirements for applying text recognition on the lensless system. To be specific, (1) size of target text in image should be sufficiently large to satisfy the resolution limitation in the lensless camera system. (2) Text features are significant and can be easily separated from the background. (3) Since deep learning method was applied for lensless image reconstruction, it would be better to cover the text features of test set in the training set, i.e., the application scenario should be simple and pre-specified.

**Table 6. Performance of Edit distance and Crwr on the IIIT5K dataset of standard images and lensless images. The effect of the image complexity was explored.**

| Mode | | Total | Edit Distance | Crwr |
|---|---|---|---|---|
| Simple | Standard | 1272 | 255 | 88.80% |
| | Lensless | 1272 | 704 | 71.78% |
| Complex | Standard | 1857 | 705 | 78.72% |
| | Lensless | 1857 | 2315 | 51.23% |

### C. Test for Variations in Optical Condition

Variations in optical condition were tested here to verify the performance of lensless imaging system in real environment. These variations contained light intensity and target's three-dimensional (3D) position. The optical conditions of NCD_40_TRAIN in Sec 4.1 were taken as a baseline, i.e., light intensity= 0% (15 lux), 3D position (x,y,z)=(0,0,0). For the test system, the coordinate system is shown in Fig.11 (a). Unit for light intensity measurement is lux and position is measured in centimeters. The model trained with only NCD_40_TRAIN in the baseline condition was used to test on NCD_40_TEST with changing the light intensity from -20% (12 lux) to +20% (18 lux), xy position from (-2,-2,0) to (+2,+2,0) and z position from (0,0,-2) to (0,0,+2).

Fig. 11 (b) shows the text recognition results when the light intensity changed. When the intensity range was between -10% and +10%, the lensless imaging system maintained relatively high accuracy. Fig. 11(c) shows the text recognition results when the target was moved in x and y directions. In this position change test, the performance of text recognition still remained a high level. However, for target moving in z direction, shown as Fig.

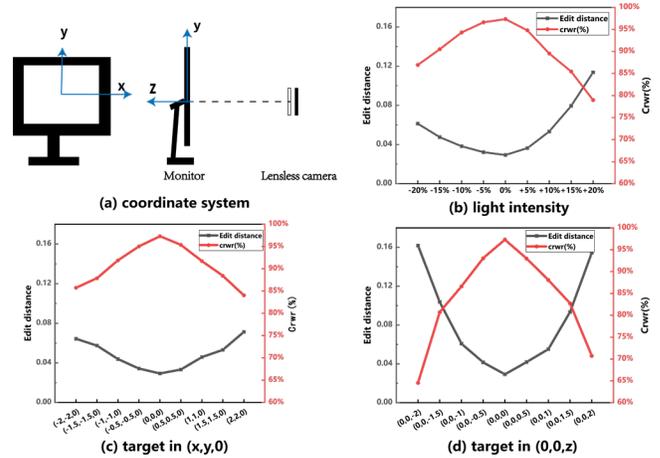

**Fig. 11.** (a) coordinate system. Result of test for variations in optical condition:(b) Crwr and versurs light intensity and average Edit distance versurs light intensity; (c) Crwr and versurs xy position and average Edit distance versurs xy position; (d) Crwr and versurs z position and average Edit distance versurs z position.

11(d), accuracy decreased dramatically when target moved away from the training position. In general, the lensless imaging system is relatively robust to target's xy directions translation and variation of light intensity, but has a weak robustness to target movement in z direction.

## 5.CONCLUSIONS

In this study, the potential and efficacy of lensless imaging systems were assessed for text detection and recognition. A framework exhibiting deep learning-based pipeline structure was developed to detect and recognize text with three steps from raw data captured by lensless cameras. First, lensless imaging model U-Net restored images from raw data. Subsequently, the text detection model CTPN extracted text regions in the reconstructed lensless images. Lastly, the text recognition model CRNN translated those image-based sequence into text. Compared with the method only focusing on the image reconstruction, U-Net in the pipeline was able to supplement the imaging details by enhancing factors related to character categories in the reconstruction process. In this way, the textual information can be more effectively detected and recognized by the CTPN and CRNN with the less artifacts and high clarity reconstructed lensless images, which further promoted the effect of lensless text detection and recognition. Moreover, the effect of the lensless text detection and recognition for scenes of different complexity was investigated by performing experiments on two distinct datasets. As verified from the results, under text with simple background (e.g., NCD), the lensless imaging system could exhibit similar performance to lens-based camera for text detection and recognition. Furthermore, with the increase in the diversity of text fonts, (e.g., IIIT5K) the effect of lensless text recognition was reduced, whereas it remained sufficient for considerable applications.

As revealed from the experiments, the performance of lensless text detection and recognition is largely dependent of several factors (e.g., image complexity, word length and text size). In brief, targets satisfying the conditions below can achieve considerable results on text recognition detection when the lensless imaging system is applied. To be specific, the background is relatively simple; the font style is in a limited set and can be pre-specified; the text length is less than a certain value, and the text size exceeds a certain value, the certain values for text length and text size are determined by the lensless imaging system. Moreover, the small number and quality of lensless images available for lensless imaging training acts as a main limiting factor. A larger dataset can help improve the performance furtherly.

Given the mentioned conclusions, lensless text detection and recognition have numerous possible application scenarios, especially for scenes with privacy protection requirements or confinement in thickness of detection devices (e.g., products tag scanning, machine internal inspections in industry). Furthermore, the work can be extended to non-textual areas, as long as the elements in the target scene have limited categories (e.g., in this project A-Z,0-9), including detection and recognition of circuit elements on lensless cameras.

**Funding.** This work was supported by the National Natural Science Foundation of China under Grant No. U19A2054.

**Disclosures.** The authors declare no conflicts of interest.

**Data availability.** Data underlying the results presented in this paper are not publicly available at this time but may be obtained from the authors upon reasonable request.

## References

1. Antipa, N., Kuo, G., Heckel, R., Mildenhall, B., Bostan, E., Ng, R., & Waller, L, "DiffuserCam: lensless single-exposure 3D imaging," Optica. 5(1), 1-9 (2018).
2. Zhou, H., Feng, H., Hu, Z., Xu, Z., Li, Q., & Chen, Y, "Lensless cameras using a mask based on almost perfect sequence through deep learning," Optics Express. 28(20), 30248-30262 (2020).
3. Bae, D., Jung, J., Baek, N., & Lee, S. A, "Lensless Imaging with an End-to-End Deep Neural Network," in 2020 IEEE International Conference on Consumer Electronics-Asia (ICCE-Asia) (2020), pp. 1-5.
4. Kim, G., Kapetanovic, S., Palmer, R., & Menon, R, "Lensless-camera based machine learning for image classification," arXiv: 1709.00408 (2017).
5. Tan, J., Niu, L., Adams, J. K., Boominathan, V., Robinson, J. T., Baraniuk, R. G, "Face detection and verification using lensless cameras," IEEE Transactions on Computational Imaging. 5(2), 180-194 (2018).
6. Asif, M. S., Ayremlou, A., Sankaranarayanan, A., Veeraraghavan, A., & Baraniuk, R. G, "Flatcam: Thin, lensless cameras using coded aperture and computation," IEEE Transactions on Computational Imaging. 3(3), 384-397 (2016).
7. Tajima, K., Shimano, T., Nakamura, Y., Sao, M., & Hoshizawa, T, "Lensless light-field imaging with multi-phased fresnel zone aperture," in IEEE International Conference on Computational Photography (2017), pp. 1-7.
8. P. R. Gill, "Odd-symmetry phase gratings produce optical nulls uniquely insensitive to wavelength and depth," Opt.Lett. 38(12), 2074–2076 (2013).
9. P. R. Gill and D. G. Stork, "Lensless ultra-miniature imagers using odd-symmetry spiral phase gratings," in Computational Optical Sensing and Imaging (Optical Society of America) (2013) paper CW4C-3.
10. Monakhova, K., Yurtsever, J., Kuo, G., Antipa, N., Yanny, K., & Waller, L, "Learned reconstructions for practical mask-based lensless imaging," Optics express. 27(20), 28075-28090 (2019).
11. S. Boyd, N. Parikh, E. Chu, B. Peleato, and J. Eckstein, "Distributed optimization and statistical learning via the alternating direction method of multipliers," Found. Trends Mach. Learn. 3, 1-122 (2010).
12. A. Sinha, J. Lee, S. Li, and G. Barbastathiset, "Lensless computational imaging through deep learning," Optica. 4 (9), 1117-1125 (2017).
13. Y. Li, Y. Xue, and L. Tian, "Deep speckle correlation: a deep learning approach toward scalable imaging through scattering media," Optica. 5(10), 1181–1190 (2018).
14. S. Li, M. Deng, J. Lee, A. Sinha, and G. Barbastathiset, "Imaging through glass diffusers using densely connected convolutional networks," Optica. 5(7), 803–813 (2018).
15. J. Zhang and B. Ghanem, "ISTA-Net: Interpretable optimization-inspired deep network for image compressive sensing," in Proceedings of the IEEE conference on Computer Vision and Pattern Recognition (2018), pp. 1828–1837.
16. Busta M, Neumann L, Matas J, "Fastext: Efficient unconstrained scene text detector," in Proceedings of the IEEE International Conference on Computer Vision (2015), pp. 1206-1214.
17. Tian, S., Pan, Y., Huang, C., Lu, S., Yu, K., & Tan, C. L, "Text flow: A unified text detection system in natural scene images," in Proceedings of the IEEE international conference on computer vision (2015), pp. 4651-4659.
18. J Jaderberg, M., Simonyan, K., Vedaldi, A., & Zisserman, A, "Reading text in the wild with convolutional neural networks," International journal of computer vision. 116(1): 1-20 (2016).
19. Gupta A, Vedaldi A, Zisserman A, "Synthetic data for text localisation in natural images," in Proceedings of the IEEE conference on computer vision and pattern recognition (2016), pp. 2315-2324.
20. Zhang, Z., Zhang, C., Shen, W., Yao, C., Liu, W., & Bai, X, "Multi-oriented text detection with fully convolutional networks" in Proceedings of the IEEE conference on computer vision and pattern recognition (2016), pp. 4159-4167.
21. Tian, Z., Huang, W., He, T., He, P., & Qiao, Y, "Detecting text in natural image with connectionist text proposal network," European conference on computer vision (2016), pp. 56-72.
22. Shi B, Bai X, Yao C, "An end-to-end trainable neural network for image-based sequence recognition and its application to scene text recognition," IEEE transactions on pattern analysis and machine intelligence. 39(11), 2298-2304 (2016).
23. Kuo, G., Antipa, N., Ng, R., & Waller, L, "DiffuserCam: diffuser-based lensless cameras," in Computational Optical Sensing and Imaging (2017), pp. CTu3B. 2.
24. Ronneberger O, Fischer P, Brox T, "U-net: Convolutional networks for biomedical image segmentation," International Conference on Medical image computing and computer-assisted intervention (2015), pp. 234-241.
25. Mishra A, Alahari K, Jawahar C V, "Scene text recognition using higher order language priors," BMVC-British Machine Vision Conference (2012).
26. Jaderberg, M., Simonyan, K., Vedaldi, A., & Zisserman, A, "Synthetic data and artificial neural networks for natural scene text recognition," arXiv: 1406.2227 (2014).
27. Shaohui Ruan, "text-detection-ctpn," (2019), https://github.com/eragonruan/text-detection-ctpn
28. Karatzas, D., Shafait, F., Uchida, S., Iwamura, M., i Bigorda, L. G., Mestre, S. R., ... & De Las Heras, L. P, "ICDAR 2013 robust reading competition," in 12th International Conference on Document Analysis and Recognition (2013), pp. 1484-1493.
29. Jieru Mei, "Convolutional Recurrent Neural Network," (2020), https://github.com/meijieru/crnn.pytorch.
30. Wang, Z. W., Vineet, V., Pittaluga, F., Sinha, S. N., & Kang, S. B., "Privacy-Preserving Action Recognition using Coded Aperture Videos." The IEEE Conference on Computer Vision and Pattern Recognition (CVPR) Workshops (2019), pp. 1-10.
31. Okawara, T., Yoshida, M., Nagahara, H., & Y Yagi, "Action Recognition from a Single Coded Image." IEEE International Conference on Computational Photography (2020), pp. 1-11.
32. Xiuxi Pan, Tomoya Nakamura, Xiao Chen, and Masahiro Yamaguchi, "Lensless inference camera: incoherent object recognition through a thin mask with LBP map generation," Opt. Express 29, 9758-9771 (2021)
33. Xiuxi Pan, Xiao Chen, Tomoya Nakamura, and Masahiro Yamaguchi, "Incoherent reconstruction-free object recognition with mask-

based lensless optics and the Transformer," Opt. Express 29, 37962-37978 (2021)
34. Xiao Chen, Tomoya Nakamura, Xiuxi Pan, Kazuyuki Tajima, Keita Yamaguchi, Takeshi Shimano, Masahiro Yamaguchi, "Resolution Improvement In FZA Lens-Less Camera By Synthesizing Images Captured With Different Mask-Sensor Distances," IEEE International Conference on Image Processing (2021), pp. 2808-2812.
35. Pan, X., Chen, X., Takeyama, S., & Yamaguchi, M, "Image reconstruction with transformer for mask-based lensless imaging,". Optics Letters, 47(7), 1843-1846 (2022).
36. M. J. Huiskes and M. S. Lew, "The MIR Flickr Retrieval Evaluation," in MIR '08: Proceedings of the 2008 ACM International Conference on Multimedia Information Retrieval, (ACM, New York, NY, USA, 2008).